\documentclass[runningheads]{llncs}
\usepackage{graphicx}
\usepackage{amsmath} 
\usepackage{float}
\usepackage{array}    
\usepackage{booktabs} 
\usepackage{multirow}
\usepackage{amsfonts}
\usepackage{hyperref}
\begin{document}
\title{YOLO-RS: Remote Sensing Enhanced Crop Detection Methods}
%
%
\author{
Linlin Xiao\inst{1} \and
Tiancong Zhang\inst{2}\and
Yutong Jia\inst{3} \and
Xinyu Nie\inst{1} \and
Mengyao Wang\inst{4} \and
Xiaohang Shao\inst{5}
}

\authorrunning{L. Xiao et al.}


\institute{
\email{2212080141@henu.edu.cn, nxy@henu.edu.cn} \and
University of Chinese Academy of Sciences, School of Physics and Optoelectronics Engineering, Hangzhou, China\\
\email{zhangtiancong23@mails.ucas.ac.cn} \and
Jilin University, College of Electronic Science and Engineering, Changchun, China\\
\email{jiayt1922@mails.jlu.edu.cn} \and
Taiyuan University of Technology, School of Computer Science and Technology, Taiyuan, China\\
\email{wangmengyao7056@link.tyut.edu.cn} \and
Henan Normal University, College of Electronic Science and Engineering, Xinxiang, China\\
\email{xhshao@htu.edu.cn}  
}
\maketitle              
\begin{abstract}
With the rapid development of remote sensing technologies, deep learning-based crop classification and health monitoring have emerged as key research areas. However, existing object detection methods often suffer from feature loss due to deep downsampling, background complexity, and limited multi-scale perception, leading to high false detection and omission rates—especially for small targets. To address these challenges, we propose \textbf{YOLO-RS}, a novel object detection model based on YOLOv11. YOLO-RS incorporates an Adaptive Hybrid Strategy (ACmix), a Bidirectional Feature Fusion mechanism, a Refined Multi-domain Multi-scale Feature Pyramid Network (RFAFPN), and a Contextual Anchor Attention (CAA) module. These components collectively enhance detection robustness in complex scenes and improve the accuracy of small object recognition while mitigating class imbalance. Experiments on the PDT crop health detection dataset and the CWC crop classification dataset demonstrate that YOLO-RS outperforms existing state-of-the-art methods, achieving 2–3\% improvements in F1-score, recall, and mean average precision (mAP), with only a moderate computational overhead of approximately 5.2 GFLOPs.

\keywords{remote sensing, object detection, smart agriculture}
\end{abstract}
\section{Introduction}
Crop management and monitoring are especially crucial as the world's population continues to rise and food demands continue to rise. Crop analysis and detection make extensive use of remote sensing technology, an efficient noncontact observation technique. By gathering data and photographs of the earth's surface, it assists agricultural managers in making precise agricultural decisions. Remote sensing imagery can be used to track crop development, disease and insect activity, soil quality, and other factors. In real-time, increasing agricultural production's sustainability and efficiency~\cite{b1}.

Crop recognition methods based on remote sensing photos have advanced significantly in recent years. To increase classification accuracy, researchers have employed a variety of deep learning methods, including transformers and convolution neural networks (CNN)~\cite{b2,b3}. The SPIEM and CIM modules were suggested by LR-FPN~\cite{b4} to improve feature fusion and raise detection accuracy in RS workloads.  Apan A. et al. Effectively identified agricultural diseases and pests using high-resolution remote sensing photos, increasing field management's response time~\cite{b5}. Furthermore, Bahrami H et al. successfully assisted the science of agricultural decision-making~\cite{b6} by integrating deep learning with multi-temporal remote sensing data to enable precise monitoring of crop growth stages.

The conventional YOLO model has two main shortcomings: 
\textbf{first}, it fails to capture the detailed features of \textbf{small targets} due to limitations in the feature extraction network; 
\textbf{second}, it cannot effectively utilize \textbf{multi-scale information} through standard feature fusion methods. 
To overcome these issues, this paper proposes an improved model named \textbf{YOLO-RS}. 
By integrating the \textbf{Contextual Anchor Attention (CAA)} mechanism, the \textbf{Bidirectional Feature Pyramid Network (BiFPN)}, and the \textbf{Adaptive Mixing Strategy (ACmix)}, 
\textbf{YOLO-RS} significantly enhances the detection performance for \textbf{small objects} in complex remote sensing scenarios.

This study suggests YOLO-RS, an enhanced small object identification model based on the most recent Yolov11, in order to address the shortcomings of the current YOLO models in remote sensing small object detection.  With the goal of overcoming the difficulties that conventional approaches encounter in the feature extraction and target recognition procedures, this module was specifically created to enhance the detection performance of small targets in complicated backgrounds.  The Context Anchor Attention Mechanism (CAA), Bidirectional Feature Fusion Network (BiFPN), and Adaptive Mixing Strategy (ACmix) are all integrated in YOLO-RS, which greatly enhances small target identification without appreciably raising computing complexity.  In the sections that follow, we will provide a detailed introduction to the YOLO-RS model's design concept and experimental findings.

The following are the main contributions we make in this work using BNN technology for traffic sign recognition tasks:
\begin{itemize}
    \item  We present the \textbf{YOLO-RS method}, a novel technique that combines the contextual anchor attention mechanism (CAA), bidirectional feature fusion network (BiFPN), and adaptive hybrid strategy (ACmix) to effectively detect small targets in remote sensing images.
    \item  By enhancing the weights of key characteristics, we provide the \textbf{Contextual Anchor Attention (CAA) module}, which increases the model's sensitivity to small and challenging-to-detect objects in remote sensing photos.
    \item  Our design achieves \textbf{remarkable performance improvements}: YOLO-RS achieves a mAP of 92. 1\% in the PDT remote sensing data set, which is 3\% better than the best existing method; YOLO-RS achieves a mAP of 96. 8\% in the CWS data set, which is 4\% better than the best existing method, while maintaining a low increase in computational complexity of only 5.2 GFLOPs.
\end{itemize}

\section{Related work}
A multifaceted research system has been created by the use of remote sensing technologies in agriculture, particularly in the areas of crop health monitoring and disease and pest identification. Precision agriculture technology based on remote sensing images can significantly increase ecological sustainability and production efficiency~\cite{b1}. However, complex background interference and multiscale feature adaption remain major technological obstacles to small item recognition.  The technical development of crop detecting techniques is methodically sorted out in this work.

\textbf{Object detection algorithms' evolution}  The two primary categories of object detection techniques based on deep learning are single-stage and two-stage.  YOLO and SSD~\cite{b7,b8} reflect the single-stage approach, which use anchor boxes for classification and regression. To maximize the effect, YOLOv2 adds dense anchor boxes and bounding boxes in contrast to YOLO~\cite{b9}.  Class.  DSSD combines low-level and high-level features by adding a deconvolution module to SSD~\cite{b7}.  The region of interest (ROL) is extracted before classification and regression in the two-stage technique.  A region proposal network (RPN) is used to drive end-to-end training driven by anchor boxes in order to achieve deep feature extraction, as opposed to R-CNN and Faster R-CNN, which used to extract candidate boxes based on a selective search~\cite{b10,b11,b12}.R-FCN employed position-sensitive score maps to enhance feature space perception~\cite{b14}, mask R-CNN introduced a mask prediction branch \cite{b13}, and cascade R-CNN resolved the prediction quality deviation using a cascade optimization mechanism that dynamically modified the IoU threshold~\cite{b15}.  RefineDet~\cite{b16} redefined the anchor box's size and position twice and improved performance by combining the benefits of the single-stage and two-stage methods. Yolov11 serves as the foundational architecture in our study, which aims to increase detection accuracy while maintaining a lightweight design.

\textbf {Improving feature fusion techniques}  The feature pyramid is the conventional method for multiscale feature processing in computer vision applications~\cite{b17}.  Nevertheless, feature pyramids and deep learning make the model more complex.  Cross-scale connections were added to convolution networks in the object detection domain to link internal feature layers of various scales in order to improve this situation~\cite{b18,b19,b20}.  Numerous studies have made important advances in the depiction of multiscale features.  Based on FPN, Liu et al. took a bottom-up approach~\cite{b21}. To enhance the feature fusion capabilities, Zhao et al. implemented a multilayer pyramid structure at the back end of the backbone network and used a U-shaped module stacking strategy~\cite{b22}.  Kong et al. achieved an adaptive feature weight allocation by dynamically aggregating multiscale contexts via a global attention mechanism~\cite{b23}.  Furthermore, by utilizing a learnable weight mechanism and bidirectional cross-scale connections, we further increase the dynamic fusion efficiency of multi-scale features in our architecture by combining BiFPN with RFAconv.

The development of techniques for \textbf{small object detection} has attracted increasing research interest in recent years. Traditional methods face several inherent challenges, such as \textbf{low resolution}, \textbf{insufficient feature representation}, and \textbf{imbalanced sample distribution}. Among the various solutions explored, \textbf{data augmentation} has emerged as a critical strategy for enhancing detection performance.

Several advanced augmentation methods for small object detection have been proposed. Yu et al.\ introduced a \textbf{scale-matching strategy} by dynamically cropping images according to target size~\cite{b24}. Kisantal et al.\ proposed a \textbf{copy-paste augmentation method} to address the issue of insufficient intersection-over-union by replicating small objects and embedding them into background regions, thereby increasing both sample diversity and quantity~\cite{b25}. Chen et al.\ extended this idea by incorporating a semantic segmentation network to select context-consistent regions for object placement~\cite{b26,b27}, and further introduced \textbf{adaptive resampling} into RRNet to address background and scale mismatches. Additionally, Zoph et al.\ broke from conventional augmentation paradigms by leveraging \textbf{reinforcement learning} to implement an adaptive selection strategy, which dynamically optimizes the augmentation process and significantly boosts model generalization~\cite{b28}.

\begin{figure}[!t]
    \centering
    \includegraphics[width=1.0\linewidth]{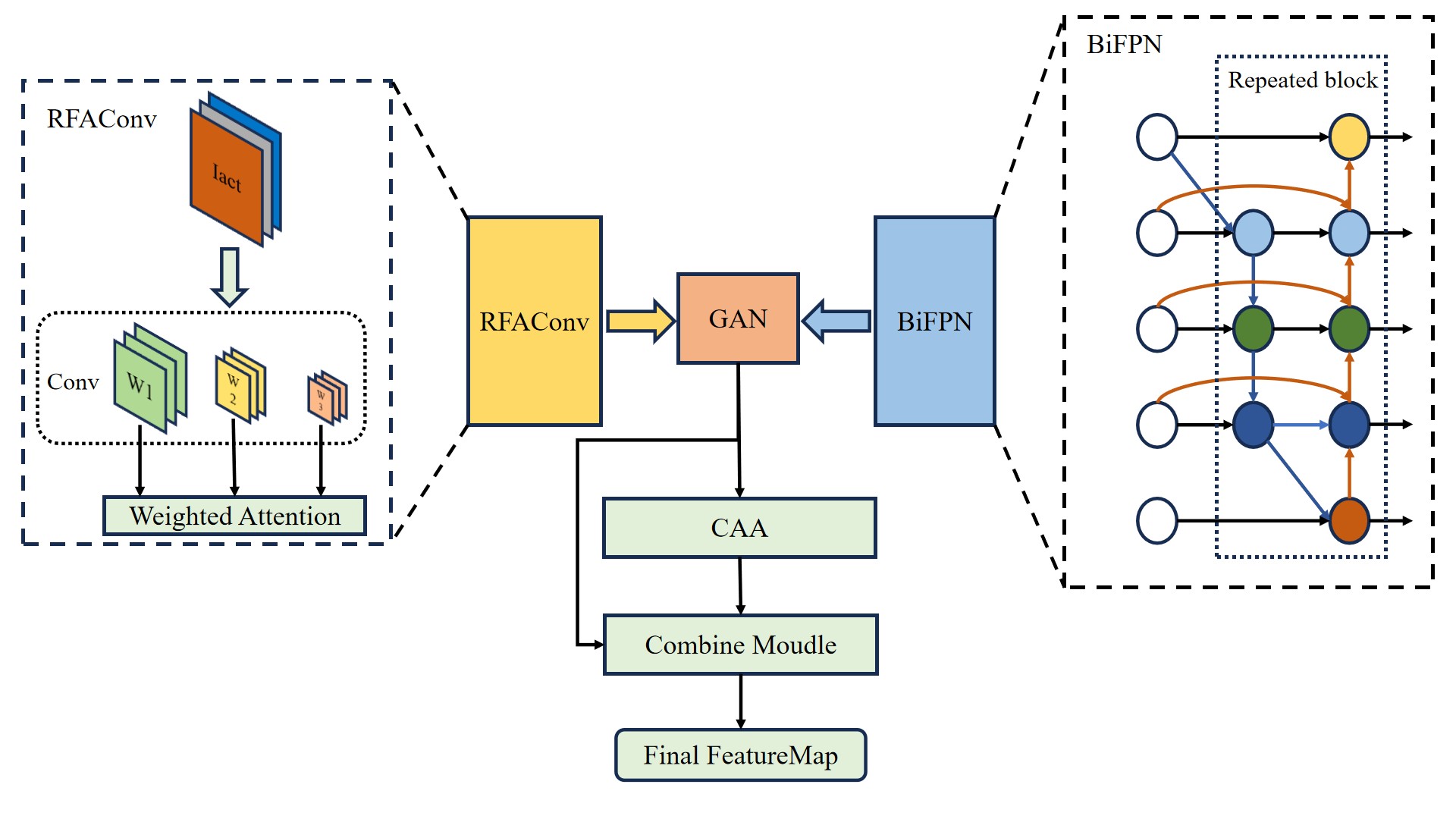}
    \caption{Diagram of the RFAFPN module framework:  Adversarial learning is applied to the output results of the RFAconv and BiFPN modules to produce output $\mathbf{F_{1}^{GAN}}$. This output is then sent to the CAA module to produce output $\mathbf{F_{2}^{CAA}}$, and $\mathbf{F_{1}^{GAN}}$ and $\mathbf{F_{2}^{CAA}}$ are combined to produce $\mathbf{F_{3}}$.}
    \label{fig:1}
\end{figure}
\section{Methodologies}
An enhanced \textbf{YOLO-RS} model is suggested in this study to improve the ability to recognize small targets in remote sensing photographs.  The model explicitly addresses the difficulties of small object detection in remote sensing images~\cite{b29,b30}. It is optimized in a number of ways based on \textbf{YOLOv11}.  The enhanced \textbf{YOLOv11} backbone network, the \textbf{Contextual Anchor Attention (CAA)} mechanism, the \textbf{Adaptive Mixing (ACmix)} module, and the \textbf{RFAFPN} are the primary elements of the \textbf{YOLO-RS} model.  The model can effectively extract and merge features in complicated backdrops with these designs, improving the properties of small objects.

\subsection{Contextual Anchor Attention Mechanism (CAA)}
 One of the YOLO-RS's global attention modules is the contextual anchor attention mechanism (CAA).  In order for the model to concentrate more on important features during feature extraction and increase the model's resilience in complex situations, it seeks to improve the model's capacity to aggregate contextual information between features at various levels.  In order to create an attention-weight map and bolster key features, CAA takes feature maps from several YOLOv11 layers and computes the feature map's global contextual information using techniques like global average pooling~\cite{b31}.  This processing technique greatly enhances small item identification performance by allowing the model to efficiently concentrate on crucial features in complex circumstances.

\subsection{Adaptive mixing (ACmix) module}
The model cannot identify several target categories during training because of the unequal frequency of occurrence of these categories in remote sensing images.  Low accuracy in detecting small targets and imbalanced categories in remote sensing photos have been resolved with the introduction of the ACmix module~\cite{b32}.  To highlight the characteristics of a few categories, the ACmix module dynamically modifies the contrast of the sample images based on the category frequency after first receiving a set of sample images and their corresponding labels from a remote sensing image dataset that includes multiple crops and backgrounds.  Lastly, fresh training samples are created by fusing the samples of a few categories with those of other categories using the sample mixing technique. This fresh collection of training samples enhances the model's detection performance on a few categories and aids in the model's more thorough learning of the features of other categories.

\subsection{Multi-sensory field multi-scale feature processing module (RFAFPN)}
To improve the extraction and fusion of small object features, the \textbf{Multi-sensory Field Multi-Scale Feature Processing (RFAFPN)} module combines the benefits of \textbf{RFAconv} and \textbf{BiFPN}.  By introducing \textbf{Generative Adversarial Network (GAN)}, \textbf{RFAconv}, and \textbf{BiFPN}, this module creates a dynamic game process to determine whether the image is real based on the discriminant network's probabilistic output, resulting in a more expressive feature map.

In adversarial learning, we generate possible optimizations and automatically decode them with the following loss in order to determine the weights of the feature encoder F (RFAconv), decoder D (BiFPN), and compressed scene representation for all training scenes:

\begin{equation}
\mathcal{L}=\lambda_{x}\cdot\mathcal{L}_{x}+\lambda_{c}\cdot\mathcal{L}_{c}+\lambda_{L1}\cdot\mathcal{L}_{L1}
\end{equation}

We set $\lambda_{x}$, $\lambda_{c}$, and $\lambda_{L1}$ to 1 for the loss weights.  These include the binary variable $y_{i}^{v}$ coordinate ${K}_{i}$ belonging to the voxel ${V}_{v}$ and the relative coordinate of the pixel point ${K}_{i}$.  The binary cross entropy for the confidence $c_{i}^{v}$, or the classification loss, is the second term:

\begin{equation}
\mathcal{L}_{x}=\sum_{v}\sum_{i}(y_{i}^{v})\cdot\left\lVert (\mathbf{K}_{i}^{v}+\mathbf{O}_{v}) - \mathbf{K}_{i}^{*} \right\rVert_{2}
\end{equation}

The third term creates a compact representation by enforcing sparsity, which is described as

\begin{equation}
\mathcal{L}_{L1}=\sum_{v}\sum_{t = 1}^{T}\lVert w_{v}^{t} \rVert_{1}
\end{equation}

The design of the \textbf{RFAconv} module is centered on resolving the issue of tiny targets' inadequate feature extraction.  In order to achieve this, \textbf{RFAconv} uses convolution kernels of different sizes, which allow it to concentrate on a wide receptive field and enhance its detection capabilities for small targets.  A weighted feature map is produced by applying an attention mechanism to each convolution output. This weighting highlights significant features and lessens the impact of noise.  In order to create an improved feature map $\mathbf{F_{1}^{GAN}}$, which serves as the basis for further feature fusion, all of the weighted feature maps are eventually spliced or added together.

The primary goal of the \textbf{BiFPN} module is to resolve the multi-scale feature fusion issue.  The detection of small targets is improved by \textbf{BiFPN}, which guarantees that features at various scales may be successfully integrated through top-down information transfer and bottom-up feature extraction.  After receiving multi-scale feature maps from \textbf{RFAconv}, \textbf{BiFPN} effectively fuses the features to produce the final fused feature maps.

An adversarial generative network is used to actualize the fusion of \textbf{RFAconv} and \textbf{BiFPN}, as illustrated in fig.\ref{fig:1}. This reflects the mutual adversarial learning process of the two.  Initially, \textbf{RFAconv}'s feature map $\mathbf{F_{1}^{GAN}}$ engages in adversarial learning with \textbf{BiFPN}'s feature map.

The context-anchored attention mechanism (\textbf{CAA}) then receives feature map $\mathbf{F_{1}^{GAN}}$ and uses the contextual information to further process the feature map, creating a new feature map $\mathbf{F_{2}^{CAA}}$.  \textbf{CAA} is able to decrease the interference of irrelevant features while strengthening important features through global contextual analysis, allowing feature map $\mathbf{F_{2}^{CAA}}$ to perform well in small object detection.

These projections are computed as $\mathbf{Q} = \mathbf{X}W_Q$, $\mathbf{K} = \mathbf{X}W_K$, and $\mathbf{V} = \mathbf{X}W_V$, where $\mathbf{X}$ is the input feature matrix, and $W_Q$, $W_K$, and $W_V$ are learnable weight matrices.
The following is $\quad A_{v} \in \mathbb{R}^{N \times d}$:

\begin{equation}
\begin{aligned}
(Q, K, V) &= A_{V}' \cdot (W_{q}, W_{k}, W_{v}) \\
Q, K &\in \mathbb{R}^{N \times d_{a}}, \quad V \in \mathbb{R}^{N \times d} \\
W_{q}, W_{k} &\in \mathbb{R}^{d \times d_{a}}, \quad W_{v} \in \mathbb{R}^{d \times d}
\end{aligned}
\end{equation}

where ${d}_{a}$ is the dimension of the query and key vectors, and ${W}_{q}$, ${W}_{k}$, and ${W}_{v}$ are shared learnable linear transformations.  The attention weights can then be determined using matrix dot products and the query and key matrices:

\begin{equation}
F = (\tilde{\alpha})_{i,j} = \text{softmax}\left(\frac{Q \cdot K^{\mathrm{T}}}{\sqrt{d_{a}}}\right). 
\end{equation}

\begin{equation}
B_{sa} = F \cdot V 
\end{equation}

The weighted sum of the value vectors with the appropriate attention weights is the output feature map $\mathbf{F_{2}^{CAA}}$.

The final feature map $\mathbf{F_{3}}$ is created by combining feature maps $\mathbf{F_{1}^{GAN}}$ and $\mathbf{F_{2}^{CAA}}$ by cross-fertilization.  Thus, the combination of \textbf{RFAconv} and \textbf{BiFPN} greatly increases the overall model detection performance and feature representation capabilities, allowing it to handle a range of obstacles in real-world applications.

\section{Experiments}
\subsection{Introduction to the dataset}
The performance of the suggested \textbf{YOLO-RS} model for small object detection in remote sensing photos is assessed in this experiment using a variety of datasets.  The PDT remote sensing dataset, the CWC dataset, and other popular remote sensing picture datasets are the main datasets used.

Crop photos taken by UAVs at high altitudes make up the PDT remote sensing dataset, which is intended to evaluate crop health.  Images of a variety of crops annotated with category information for various crops are included in the CWC dataset, which focuses on crop category determination.  To further enhance the context of model evaluation, we also make reference to a few additional widely used remote sensing image datasets, including the VHR-10 Dataset and the UC Merced Land Use Dataset~\cite{b33}.  Combining these datasets allows us to thoroughly assess the model's performance in a range of crop detection tasks, such as category judgment and health detection.

\subsection{Evaluation indicators}
We employ a number of assessment criteria, such as accuracy, precision, recall, F1-score, and mean average precision (mAP), to assess the model's performance.  These metrics can accurately depict how well the model performs in object detection.  The percentage of all predictions where the model is accurate is expressed as the accuracy ratio, which is the number of correctly classified samples divided by the total number of samples.  Conversely, precision rate (7) is the percentage of samples that are positive based on the model's prediction.  The model's ability to catch samples in the positive category is demonstrated by recall (8), which shows the percentage of all real positive samples that the model correctly predicts. When there is a category imbalance, the F1-score (9)—the reconciled average of precision and recall—is applicable.

\begin{equation}
\text{Precision} = \frac{TP}{FP + TP}
\end{equation}

\begin{equation}
\text{Recall} = \frac{TP}{FN + TP}
\end{equation}

\begin{equation}
\text{F1} = 2 \cdot \frac{\text{Precision} + \text{Recall}}{\text{Precision} \cdot \text{Recall}}
\end{equation}

By computing and averaging the AP for each category, mAP is able to more effectively synthesize the model's performance in multi-category detection, which is a crucial metric in the object detection job.  We are able to measure the model's detection performance and give data support for further model optimization by applying these evaluation criteria in a methodical manner.

\subsection{Experimental configuration}
To guarantee effective model training and evaluation, we configured the hardware and software conditions for the experiment.  The deep learning framework PyTorch 1.10, CUDA version 11.3, and Python version 3.8 were used in this experiment, which was carried out using the Ubuntu 20.04 LTS operating system.
 
The experiment's hardware setup consists of an NVIDIA GeForce RTX 3080 GPU with 10GB of video RAM that can handle training.  The batch size is set to 50, and the initial learning rate is set to 0.001 in terms of hyper-parameter settings.  SGD is chosen by the optimizer, the learning rate decay technique is used, and 100 training cycles are allotted.  Additionally, we employ a number of data improvement strategies during the training phase, such as random cropping, scaling, rotating flips, etc., to increase the model's capacity for generalization.  These improvements successfully broaden the variety of training samples and strengthen the model's resilience.

\subsection{Main experiments}
\begin{figure}[!t]
    \centering
    \includegraphics[width=1.0\linewidth]{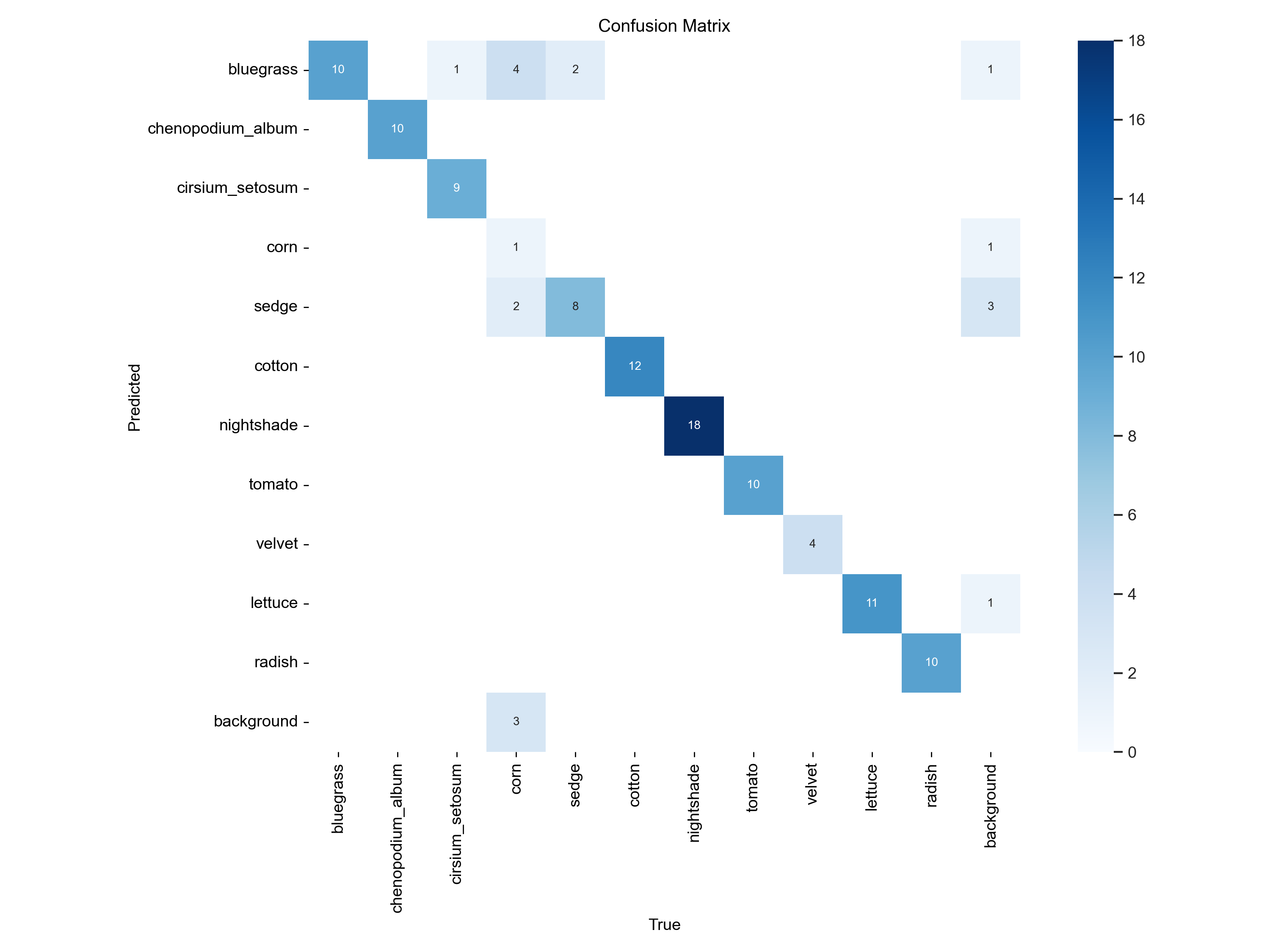}
    \caption{Confusion Matrix Normal training plot based on PDT and CWC datasets under the YOLO-RS}
    \label{fig:2}
\end{figure}

The comparison studies are designed to evaluate the relative superiority of our suggested YOLO-RS model by contrasting its performance with that of other cutting-edge object identification methods.  We proposed the YOLO-RS model for small object detection and compared it with the SOTA models EfficientDet-D7 and Cascade R-CNN in the same type of tasks in different scenarios. They are both used for high-precision small target detection.  These models can give us a useful foundation for comparison and are well-regarded in the object detection community.  To guarantee experiment fairness, all comparison models are trained using the same hyperparameter values and on the same training set.  We illustrate the experimental results using fig.\ref{fig:2}.

We documented each model's performance data on the same validation set.  In terms of accuracy, precision, recall, and mAP, our enhanced model performs better than the other comparison models, as demonstrated by the experimental findings (see Table~\ref{tab:controlled_pdt}), which are 90.1\% for YOLOv7 and 72.3\% for EfficientDet~\cite{b34}.  Additionally, YOLO-DP's mAP is 94.5\%.  On the other hand, with a mAP of 89.0\%, our model performs better than the others in the majority of the criteria. These experimental findings clearly show the benefits of the suggested model in the task of identifying small targets in remote sensing photos, particularly when it comes to recognizing small targets and complicated backdrops, which shows greater efficacy and practicality.  These findings set the groundwork for future research initiatives in addition to offering strong technical backing for real-world uses like agricultural surveillance.

\begin{table}[!t]
\centering
\caption{Main result on the PDT dataset and  the CWC dataset}
\label{tab:controlled_pdt}
\begin{tabular}{l l c c c c c c}
\hline
Datasets & Approach & P (\%) & R (\%) & mAP@.5 (\%) & mAP@.5-.95 (\%) & F1 & Gflops \\
\hline
\multirow{14}{*}{PDT (LL)} & SSD & 84.5 & 87.7 & 85.1 & -- & 0.86 & 273.6 \\
 & EfficientDet & 92.6 & 73.4 & 72.3 & -- & 0.82 & 11.5 \\
 & RetinaNet & 93.3 & 65.3 & 64.2 & -- & 0.79 & 109.7 \\
 & CenterNet & 95.2 & 67.4 & 66.5 & -- & 0.79 & 109.7 \\
 & Faster-RCNN & 57.8 & 70.5 & 61.7 & -- & 0.64 & 401.7 \\
 & YOLOv3 & 88.5 & 88.1 & 93.4 & 65.7 & 0.88 & 155.3 \\
 & YOLOv5s & 88.8 & 88.8 & 94.7 & 66.1 & 0.88 & 20.8 \\
 & YOLOv5s\_7.0 & 88.9 & 88.2 & 94.2 & 67.0 & 0.89 & 16.0 \\
 & YOLOv6s & -- & -- & 91.4 & 63.2 & -- & 40.1 \\
 & YOLOv7 & 87.4 & 82.6 & 90.1 & 55.5 & 0.85 & 105.1 \\
 & YOLOv8s & 88.7 & 87.5 & 94.0 & 67.9 & 0.88 & 28.6 \\
 & WeedNet-R & 87.7 & 48.1 & 70.4 & -- & 0.62 & 19.0 \\
 & YOLO-DP & 90.2 & 88.0 & 94.5 & 67.5 & 0.89 & 11.7 \\
 & YOLOv11 & 86.5 & 86.1 & 92.9 & 65.6 & 0.86 & 6.3 \\
 & YOLO-RS & 92.1 & 89.0 & 89.7 & 68.1 & 0.90 & 11.5 \\
\midrule
 \multirow{14}{*}{CWC  (LL)} & SSD & 84.5 & 87.7 & 85.1 & -- & 0.86 & 273.6 \\
 & EfficientDet & 97.2 & 98.6 & 98.6 & -- & 0.90 & 11.5 \\
 & RetinaNet & 95.1 & 98.3 & 98.0 & -- & 0.97 & 261.3 \\
 & YOLOv3 & 86.8 & 89.4 & 93.4 & 82.3 & 0.88 & 154.7 \\
 & YOLOv4s & 87.3 & 87.9 & 91.9 & 81.5 & 0.88 & 20.8 \\
 & YOLOv5s\_7.0 & 88.6 & 88.7 & 93.0 & 81.2 & 0.89 & 16.0 \\
 & YOLOv6s & -- & -- & 92.7 & 84.3 & -- & 68.9 \\
 & YOLOv7 & 93.1 & 76.4 & 88.1 & 75.6 & 0.84 & 105.1 \\
 & YOLOv8s & 92.0 & 89.1 & 94.0 & 86.2 & 0.91 & 28.6 \\
 & WeedNet-R & 86.1 & 51.8 & 71.6 & -- & 0.65 & 19.0 \\
 & YOLO-DP & 92.9 & 87.5 & 91.8 & 81.0 & 0.90 & 11.5 \\
 & YOLOv11 & 92.3 & 89.2 & 93.1 & 81.7 & 0.89 & 6.3 \\
 & YOLO-RS & 89.6 & 90.1 & 96.8 & 91.0 & 0.90 & 11.5 \\
\hline
\end{tabular}
\end{table}

\subsection{Ablation experiments}
The purpose of ablation experiments is to assess the distinct ways in which various modules affect model performance.  We confirmed the efficacy of each module by observing changes in model performance when component modules were removed.

To serve as a foundation for further comparisons, the full model YOLO-RS is first provided; after that, the BiFPN module is taken out of the entire model.  The significance of this module in feature fusion is demonstrated by the experimental results, which demonstrate that the accuracy and mAP of the model both decline after BiFPN is removed.  The RFAconv module is then eliminated from the entire model. By adding the perceptual field attention mechanism, RFAconv improves the model's capacity to concentrate on features of various scales.  According to the experimental findings, removing this module lowers small target identification ability, particularly in complicated backgrounds.  The CAA module is then eliminated from the entire model.  Following this experiment, the model's performance on the tiny item identification task likewise declines, confirming the CAA module's efficacy.  Lastly, in order to restrict sample diversity and model robustness, we eliminate the ACmix module from the entire model.

Following the removal of this collection of modules from the PDT dataset, we found that the combination of each module significantly improved the results, as indicated in Tables~\ref{tab:ablation_study_2} and ~\ref{tab:ablation_study_3}.  The outcomes of this set of ablation tests show how each module contributes differently to enhancing model performance and serve as a valuable guide for the construction of future models.

\begin{table}[!t]
\centering
\caption{Ablation study of RFAFPN on PDT dataset (LL)}
\label{tab:ablation_study_2}
\begin{tabular}{l c c c c c}
\hline
\textbf{Approach} & \textbf{P (\%)} & \textbf{R (\%)} & \textbf{mAP@.5 (\%)} & \textbf{F1} & \textbf{Gflops} \\ 
\hline
 RFAFPN & 89.6 & 89.0 & 92.1 & 0.89 & 11.5 \\ 
 w/o RFAconv & 85.3 & 72.6 & 76.3 & 0.78 & 8.7 \\ 
 w/o BiFPN & 88.3 & 84.4 & 86.8 & 0.86 & 10.1 \\ 
 w/o Self-attention & 80.0 & 70.0 & 75.0 & 0.74 & 7.5 \\ 
\hline 
\end{tabular}
\end{table}

\begin{table}[!t]
\centering
\caption{Ablation study of YOLO-RS on PDT dataset (LL)}
\label{tab:ablation_study_3}
\begin{tabular}{l c c c c c}
\hline
\textbf{Approach} & \textbf{P (\%)} & \textbf{R (\%)} & \textbf{mAP@.5 (\%)} & \textbf{F1} & \textbf{Gflops} \\ 
\hline
 Yolov-RS & 89.6 & 89.0 & 92.1 & 0.89 & 11.5 \\ 
 w/o RFAFPN & 85.3 & 72.5 & 76.3 & 0.78 & 8.7 \\ 
 w/o CAA & 88.3 & 84.4 & 86.8 & 0.86 & 10.1 \\ 
 w/o ACmix & 92.7 & 72.3 & 70.5 & 0.81 & 10.3 \\ 
\hline 
\end{tabular}
\end{table}

\subsection{Visualizing labels and predictions}
\begin{figure}[!t]
    \centering
    \includegraphics[width=0.8\linewidth]{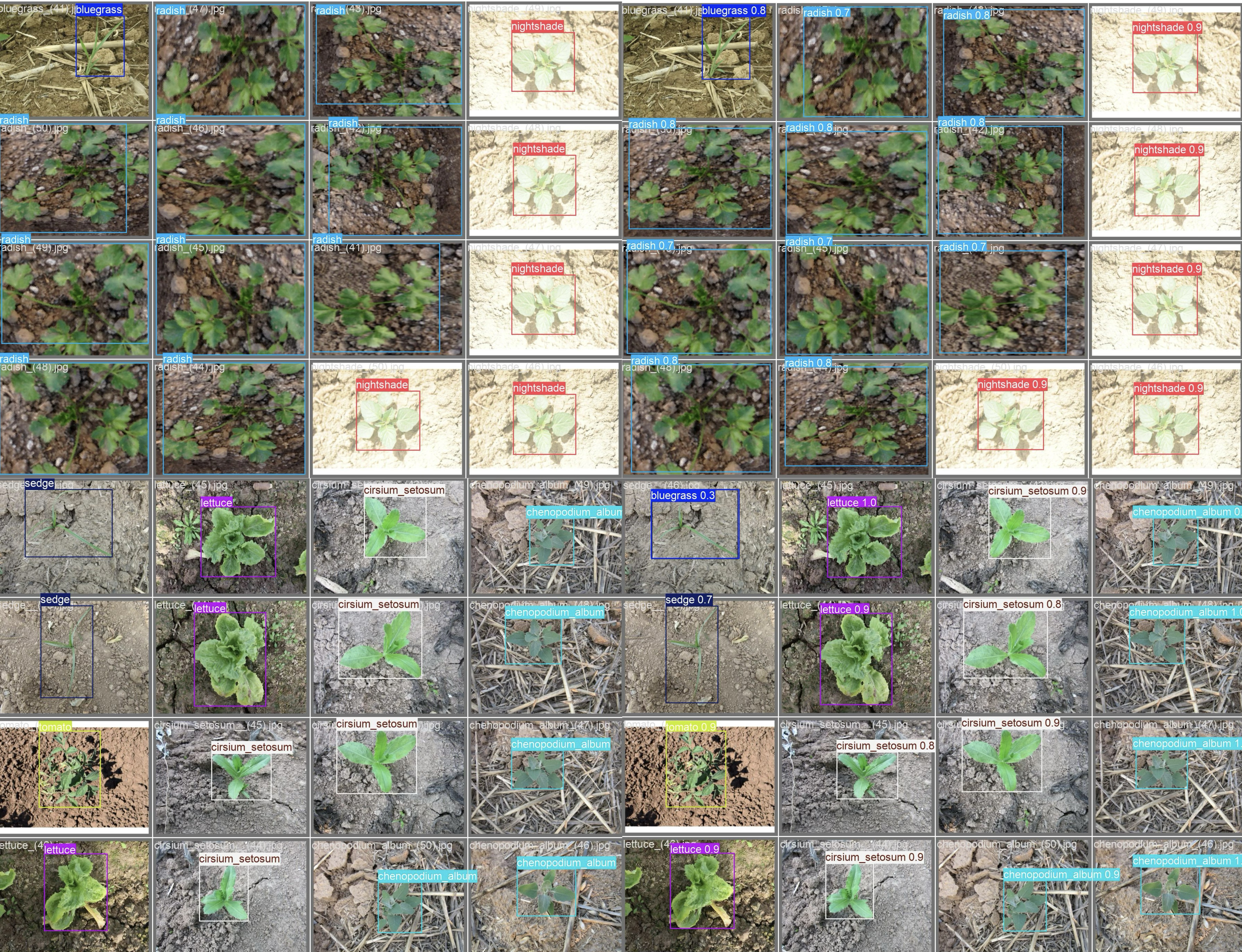}
    \caption{Visualizing Diversity Estimates Our Method}
    \label{fig:3}
\end{figure}

The YOLO-RS architecture is intuitively illustrated in Fig.~\ref{fig:3}, showcasing its application across various crop identification tasks, such as health detection and category classification, using the PDT and CWC datasets. The model's recognition performance corresponds closely with the quantitative results presented in Table~\ref{tab:controlled_pdt}. The figure includes both ground-truth labels and model predictions, providing a visual comparison to assess the accuracy of the detection results. This visualization highlights the clear advantages of the YOLO-RS architecture in small object detection tasks within remote sensing imagery.

\section{Conclusion}
The new object detection model YOLO-RS is presented in this paper.  Small target feature expression is greatly enhanced by the model's integration of three modules: the contextual anchor attention mechanism (CAA), adaptive hybrid strategy (ACmix), and multi-receptive field feature fusion network (RFAFPN). Through the spatial channel collaborative attention mechanism, the CAA module improves the response of key areas; the multi-scale fusion network (RFAFPN) optimizes the crop micromorphology analysis ability through cross-layer feature interaction; and ACmix dynamically balances the category weights to alleviate the issue of sample imbalance. Experiments on the CWS classification dataset and the PDT remote sensing crop health detection dataset reveal that the mAP of YOLO-RS achieves 96.8\% and 92.1\%, respectively, which is 3\%–4\% higher than the current best technique.  While the computational complexity is only increased by 5.2 GFLOPs, the accuracy and recall rate are simultaneously improved, striking a compromise between efficiency and performance.  In order to enhance environmental adaptation, future research will concentrate on combining multi-modal data (such as heat infrared and SAR pictures).  Simultaneously, it lessens reliance on labeled data and encourages the low-cost use of this technology in precision agriculture when paired with transfer learning.

%
%
%
%


\bibliographystyle{IEEEtran}
 
\bibliography{main.bib}
 
\end{document}